\documentclass[final]{cvpr}

\usepackage{times}
\usepackage{epsfig}
\usepackage{graphicx}
\usepackage{amsmath}
\usepackage{amssymb}
\usepackage{color}

\DeclareMathOperator*{\argmax}{arg\,max}

\usepackage[pagebackref=true,breaklinks=true,colorlinks,bookmarks=false]{hyperref}

\begin{document}

\title{Learning Dynamics via Graph Neural Networks for \\
Human Pose Estimation and Tracking}


\author{
Yiding Yang$^1$\footnotemark[1],
Zhou Ren$^2$,
Haoxiang Li$^2$,
Chunluan Zhou$^2$,
Xinchao Wang$^{1,3}$\footnotemark[2],
Gang Hua$^2$\\ %
$^1$Stevens Institute of Technology,
$^2$Wormpex AI Research,
$^3$National University of Singapore\\
{\tt\small
\{yyang99, hli18, xinchao.wang\}@stevens.edu,
renzhou200622@gmail.com,}\\
{\tt\small
czhou002@e.ntu.edu.sg,
ganghua@gmail.com
}
}

\maketitle

\thispagestyle{empty}
\pagestyle{empty}

\footnotetext[1]{The work is partially done when the author is an internship at
Wormpex AI Research.}
\footnotetext[2]{Corresponding author.}

\begin{abstract}
Multi-person pose estimation and tracking 
serve as crucial steps for video understanding.
Most state-of-the-art approaches 
rely on first estimating 
poses in each frame and only then
implementing data association
and refinement.
Despite the promising results achieved,
such a strategy is inevitably
prone to missed detections
especially in heavily-cluttered scenes,
since this tracking-by-detection
paradigm is, by nature, largely dependent
on visual evidences that are absent
in the case of occlusion. 
In this paper, we propose a 
novel online approach to
learning the pose dynamics,
which are independent of 
pose detections in current fame,
and hence may serve as a robust
estimation even in challenging scenarios
including occlusion.
Specifically, we derive this 
prediction of dynamics through 
a graph neural network~(GNN)
that explicitly accounts for
both spatial-temporal and visual information.
It takes as input the historical pose tracklets and 
directly predicts the corresponding poses 
in the following frame for each tracklet. 
The predicted poses will then
be aggregated with the 
detected poses, if any, at the same frame
so as to produce the final pose,
potentially recovering the occluded
joints missed
by the estimator. 
Experiments on 
PoseTrack 2017 and PoseTrack 2018 datasets
demonstrate that
the proposed 
method achieves results superior to
the state of the art 
on both human pose estimation and tracking tasks. 

\end{abstract}
\vspace{-1em}

\section{Introduction}
\begin{figure}[t]
\begin{center}
    \includegraphics[width=0.98\linewidth]{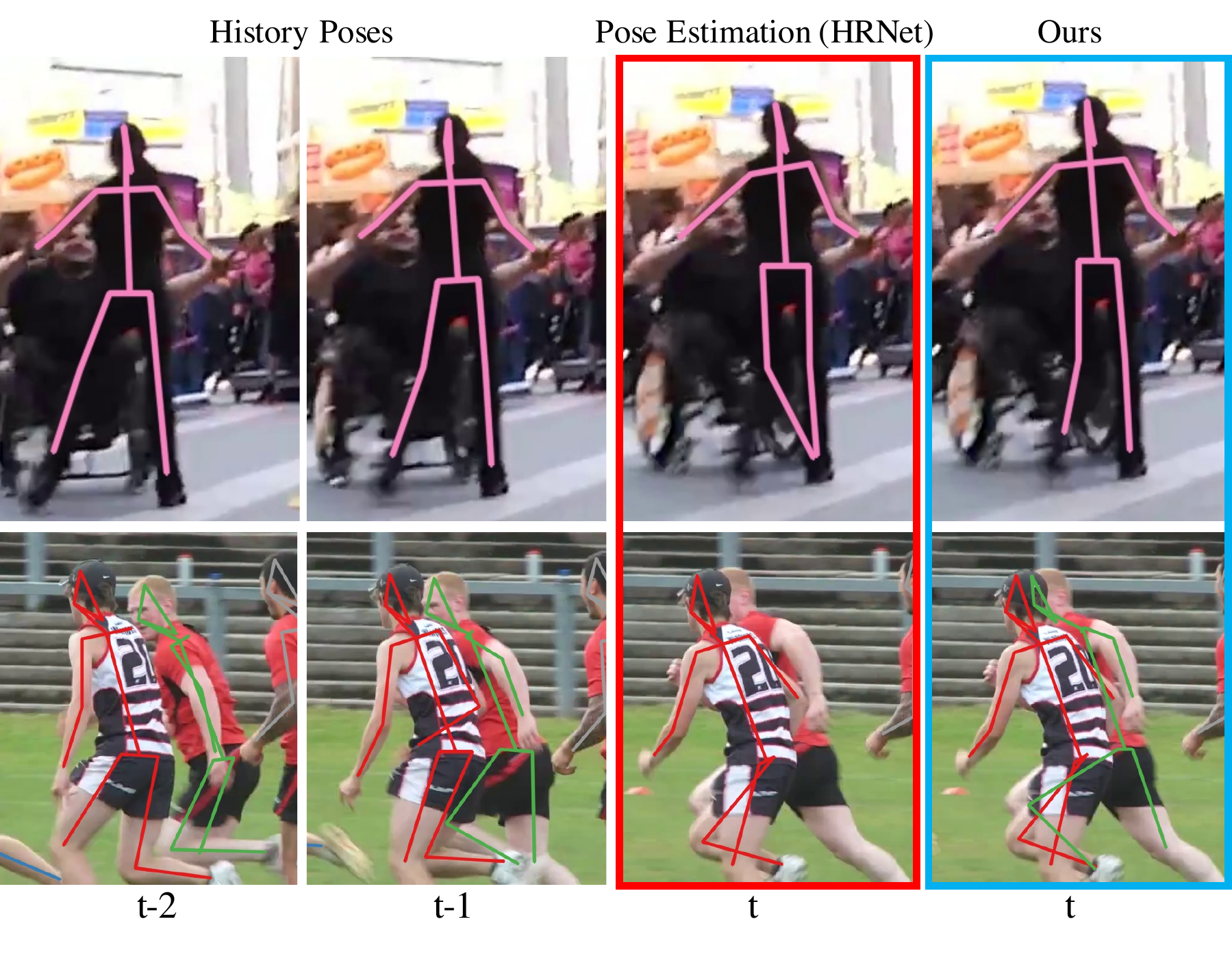}
\end{center}
   \vspace{-1.5em}
   \caption{
    By modeling the pose dynamics 
    from history poses 
    through a graph neural network,
    our method learns a pose prediction
    that is robust to challenging scenes,
    such as motion blur~(top)
    and occlusion~(bottom).
    In both cases,
    the visual-based 
    HRNet~\cite{sun2019deepHRNet}
    fails to locate the joints,
    yet our approach delivers
    dependable pose estimations.  
   }
   \vspace{-1.8em}
\label{fig:head_example}
\end{figure}

Multi-person pose estimation and tracking
find their applications in a wide spectrum
of scenarios including behavior analysis
and action recognition,
and have therefore received increasing attention
in recent years~\cite{FlowTrack,STAF,BUTD}.
Despite often coupled together,
they focus on slightly different aspects:
the former 
aims to locate human joints
in each frame of an input video,
while the latter one aims to 
associate joints that belong
to the same human across frames.
It has been long considered as
a challenging task
due to various factors,
including but not limited to
camera motions,
complex backgrounds, 
and mutual occlusions.

Thanks to the recent advances
of deep learning techniques,
pose estimation and tracking have
witnessed unprecedented results
in the past years.
Existing methods can be broadly
categorized into
two streams,
bottom-up methods~\cite{STAF,BUTD,RPAF,ST-Embed}
and top-down methods~\cite{FlowTrack,TKMRNet,CombDet}.
Bottom-up methods first generate joint candidates and
then group the joints into a person detection.
The grouped joints are then associated across frames to generate
the final pose tracking results. 
Top-down methods,
on the other hand, 
first detect human candidates in a single frame and
then estimate the human poses for each candidate. 
The estimated human poses are 
associated across frames to achieve pose tracking. 
Methods from both streams
have produced promising results
on various scenarios~\cite{TKMRNet,CombDet}.

In spite of the encouraging results,
state-of-the-art pose estimation
and tracking approaches 
remain prone to missed detections
especially in highly-cluttered 
and fast-motion scenes. 
This is not totally unexpected,
since by nature they rely on
first detecting either
joints or human bodies in
a scene using a visual-based 
detector, and only then
carrying out data association
to link the detections into tracks.
In challenging scenarios
such as crowded or blurred
scenes, the joint- or human-detector
would inevitably fail
due to the absent image evidences.
Although some succeeding
refinement steps would
mildly remedy the flawed
estimations, they are
are still largely dependent on visual cues
and hence incompetent
to fully tackle missed detections.

We propose in this paper 
a novel approach by explicitly
looking into the \emph{dynamics} 
of human poses within image sequences.
In contrast to state-of-the-art approaches
that rely on first 
detecting human or joints in each frame,
which is again prone to failures
in the absence of detection evidences,
our approach first \emph{predicts}
poses in a frame 
from a track of history
without looking at
any detection cue.
This strategy allows 
us to free our dependency on 
the detection evidences
and consequently produce
a legitimate state of human pose
at the very first place.
Specifically,
in our approach this prediction
step is accomplished
through a graph neural network~(GNN)
that takes as input
a track of history poses
in previous frames. 
Next, the predicted pose 
is aggregated with the 
detected poses, if any,
in the same frame
to produce the final pose,
in which way both 
dynamical and visual 
information are 
exploited. 
At a conceptual level,
our approach 
follows a similar spirit of
Bayesian filters, expect that
in our approach 
all parameters and features
are learned end to end.
A qualitative example is 
shown in Figure~\ref{fig:head_example},
where our dynamic-based
approach yields dependable 
pose estimation results
in the cases of motion blur
and occlusion.

Apart from the strength 
of recovering missed poses
from predictions,
the proposed approach also 
enjoys other merits. 
First, 
prior approaches
match poses between
two \emph{consecutive} frames,
which is brittle to
identify switches 
due to factors such as 
intersection of poses and fast motion.
Our approach, by contrast,
aggregates poses
within the \emph{same frame},
thanks to our prediction-based nature,
allowing us to significantly reduce the
mismatched rate.
Second, 
as compared to state-of-the-art methods,
our approach tackles pose tracking
from an additional perspective, \emph{i.e}\onedot
the motion dynamics,
which complements the visual cues
that are in many cases absent, 
resulting in gratifying final poses.

We evaluate the effectiveness 
of the proposed method on
two widely used benchmark datasets,
PoseTrack 2017 and PoseTrack 2018.
Empirical evaluations showcase that our method
outperforms state-of-the-art approaches
by a considerably large margin
on both pose estimation and tracking 
tasks. 
We also provide extensive 
analyses on the impact 
of each component in the proposed method,
and demonstrate the superiority 
of learning pose dynamics using our method.

\section{Related Work}
We briefly review the following three related topics,
including
single-frame human pose estimation,
human pose tracking,
and graph neural networks.

\subsection{Single-Frame Human Pose Estimation}
Human pose estimation methods from single images can be 
generally categorized into top-down methods and bottom-up methods. 
Bottom-up methods~\cite{cao2017realtime_bottomup,newell2017associative_bottomup,pishchulin2016deepcut_bottomup,insafutdinov2016deepercut_bottomup,cheng2020higherhrnet_bottomup} do not 
rely on human detectors. 
These methods first detect all the body joints and
then group them to form human poses.
The major challenges are robustly detecting joints in 
complex situations~(e.g. various scales, poses and cluttered background)
and correctly grouping joints from different persons particularly 
in crowds with heavy occlusions.

Top-down methods first detect the human
bounding boxes from an image and then estimate the human pose within each bounding box. 
Most top-down methods adopt off-the-shelf 
human detectors~\cite{ren2016faster,chen2019hybrid_detector,zhu2019deformable_detector}
and focus on designing efficient human pose
estimators~\cite{sun2019deepHRNet,pfister2015flowing_poseestimator}.
Pose estimation is confined for a single person
within a small area at a fixed scale. With a reliable human detector, the top-down methods can achieve accurate human pose estimation.



\begin{figure*}[t]
\begin{center}
\scalebox{0.9}{
    \includegraphics[width=0.93\linewidth]{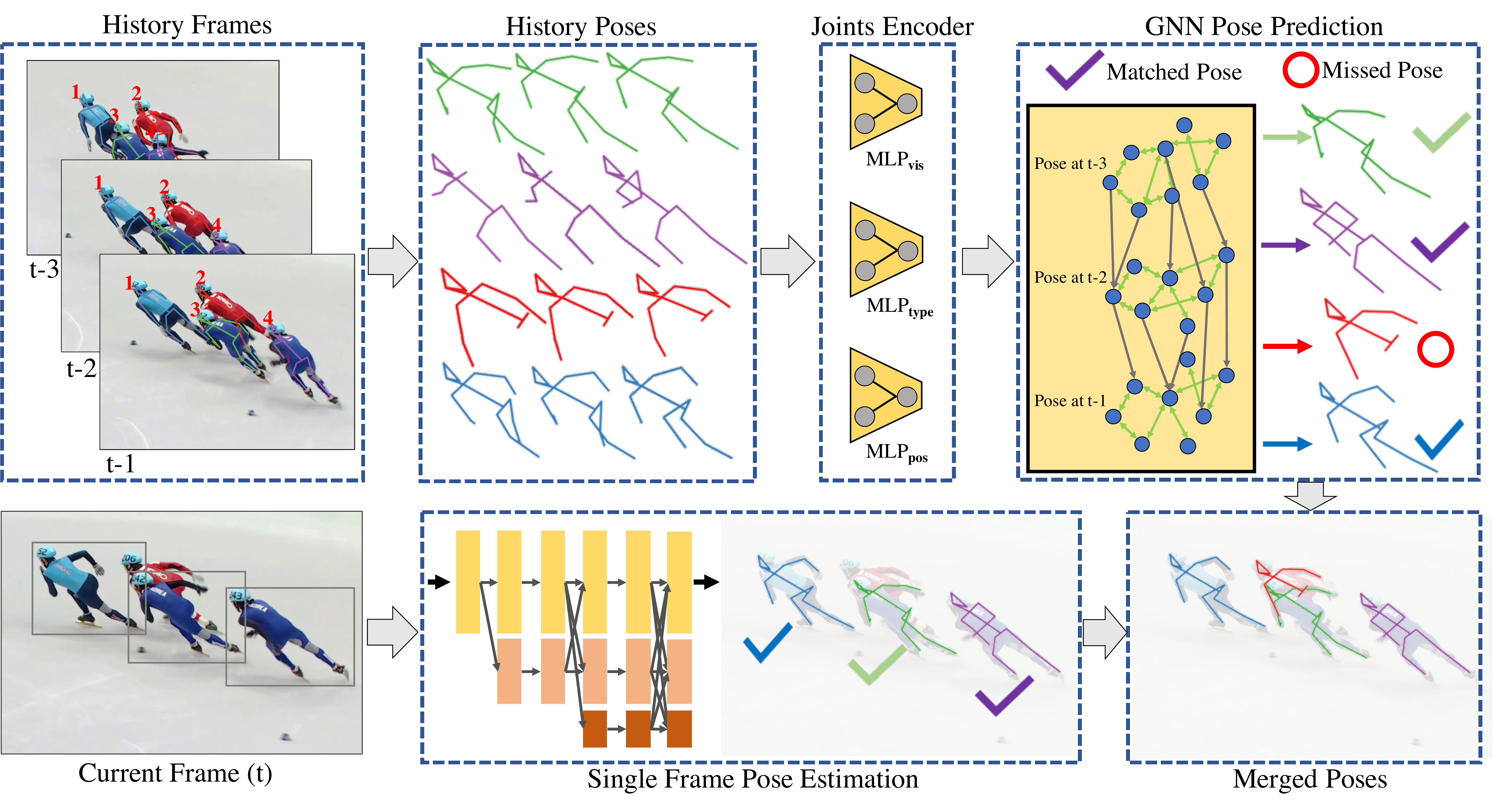}
    }
\end{center}
  \vspace{-1.5em}
  \caption{Overall pipeline of the proposed method. 
  Given the history of poses and the current frame,
  the GNN model predicts poses for each tracklet 
  in the history memory. 
  The predicted poses are then matched and merged with 
  the estimated poses to obtain the final poses in
  the current frame.}
  \vspace{-1.5em}
\label{fig:pipeline}
\end{figure*}

\subsection{Human Pose Tracking}

Extending the pose estimation to video lead to 
the human pose tracking problem, where the human poses
are estimated for each frame and
associated across frames. As a result, pose tracking
is often tackled together with human-location tracking~\cite{Wang2014_ECCV,Wang2016_TPAMI,Maksai2016_CVPR,Maksai2017_ICCV,Lan2020_IJCV}.

Bottom-up methods~\cite{STAF,ST-Embed,WangJue_ICCV_2019} in pose tracking associated 
the joints spatially and temporally without detecting human bounding boxes. 
For example, Raaj~\etal~\cite{STAF} extended the Part Affinity Field~(PAF)~\cite{cao2017realtime_bottomup} designed for single image 
pose estimation to include temporal modeling for pose tracking.
Jin~\etal~\cite{ST-Embed} proposed ST-Embed to learn the
Spatial-Temporal Embedding of joints based on
the idea of Associative Embedding~\cite{newell2017associative_bottomup}.
Both methods only model relationships of joints between two frames.

Top-down methods focus on improving single-frame pose estimation 
by exploiting temporal context and associating the estimated 
poses into human pose tracklets.
In the simple baseline method~\cite{FlowTrack},
the estimated human poses are associated by the similarity computed based on the 
optical flow between consecutive frames.
Detect-and-Track~(DAT)~\cite{DAT} utilizes
a 3D Mask R-CNN model to detect persons with key-points
from a video clip and then associates them
by comparing the locations of person detections.
CombDet~\cite{CombDet} extends a 3D network as the backbone for pose estimation
to generate a tube
of poses by directly propagating a bounding box to the neighboring
frames. KeyTrack~\cite{KeyTrack} associates the estimated human poses
pose similarities. TKMRNet~\cite{TKMRNet} matches
human poses by learning appearance embeddings of joints
and refines joints by exploiting temporal context from tracked poses.

Although some of the prior methods utilize multiple consecutive frames 
to help improve pose estimation and tracking, none of them explicitly 
model the spatial-temporal and visual dynamics of human joints. Our method 
models the pose tracking process with a Graph Neural Networks 
to learn the dynamics across frames from data.

\subsection{Graph Neural Networks}

Graph Neural Networks~(GNNs) was first developed for graph analysis such as 
node classification~\cite{kipf2016semi} 
and link prediction~\cite{zhang2018link}. 
It shows great potential in dealing with non-grid data~\cite{gilmer2017neural,ijcai2019spagan,yang2020factorizable}
and has been applied to process point clouds and
images~\cite{SplineCNN,dgcnn,mena2017sinkhorn,yang2020distilling,qiu2020hallucinating}.
For example, DGMPN~\cite{zhang2020dynamic} 
utilize GNN to capture the long range dependence among pixels in images
to enhance the feature representation.

GNN has been used to model human poses for pose-based action recognition~\cite{liu2020disentangling,cheng2020skeleton,si2019attention}
and single-frame pose estimation~\cite{wang2020global,bin2020structure}. 
For example, DGCN~\cite{qiu2020dgcn} adopts several learnt graphs to
model the relations of different joints and
propagates among them to obtain the enhanced joint feature
for better human pose estimation.

There are prior works that use GNNs for generic object tracking~\cite{gao2019graph,bao2020poseGuided}. 
Gao~\etal~\cite{gao2019graph} proposed to divide an object into several parts and learn a spatial-temporal template of the object for tracking. Bao~\etal~\cite{bao2020poseGuided} utilized GNN in their pose tracking method to exploit human structural relations to help associate human poses across frames. 
This method relies on a strong human detector as well as a strong pose estimator to generate human poses for association. 

In this paper, we propose a GNN-based predictor to estimate a potential configuration for each human pose tracklet frame by frame via leveraging the tracked pose history. 
The learnable predictor naturally models the pose tracking process 
and captures the dynamics of pose tracklets across video frames. 
Our proposed framework is capable of predicting the poses of missed human detections, which makes it robust to heavy occlusions and motion blur.


\section{Method}

Figure~\ref{fig:pipeline} shows the overall pipeline of the proposed method.
For each incoming frame, two sets of poses are computed
separately by the single-frame pose estimation module and
the GNN-based pose prediction module.
These two sets of poses are matched and merged together to generate
the final human poses for the current frame.
We introduce each components of the proposed method in the following sections.

\subsection{Single-Frame Pose Estimation}

We follow the standard pipeline of recent top-down pose trackers~\cite{FlowTrack,CombDet,TKMRNet} to perform pose estimation for each frame.
Each human detection in a frame is first cropped
and rescaled to a fixed size~(e.g. $384 \times 288$ when
HRNet is used as the backbone of human pose estimation).
The human pose estimator takes the scaled image as input
and outputs a set of feature maps
as well as a set of heatmaps $\mathbf{H}$.
The size of the generated heatmaps is typically smaller than the
input image~(e.g. $96 \times 72$ with
HRNet as the backbone).
The number of heatmaps is set to be
the number of joints, which is 15 on PoseTrack 2017
and PoseTrack 2018 datasets.
Let $\mathbf{H}_{ijk}$ be the value 
at the $(i, j)$ location of the $k$-th heatmap. 
The position of the $k$-th joint can be computed as
\begin{equation}
    l_k^* = \argmax_{(i,j)} \mathbf{H}_{ijk},
\end{equation}
where $l_k^*$ is the position within the heatmap
and can be transformed to the position in the frame
according to the center and scale information of the 
cropped image.

The training loss of the single-frame 
pose estimation model is computed against the heatmaps. 
A cropped human example is first scaled to a 
fixed size and the corresponding ground-truth joints are properly transformed to the coordinates in heatmaps. 
Let $l_k$ be the ground-truth location of the $k$-th joint in the heatmap. 
The ground truth heatmap is generated following
a 2D Gaussian distribution: 
$\mathbf{H}_{ijk}^{gt} = \exp(-\frac{||(i,j)-l_k||_2^2}{\sigma^2})$.
$\sigma$ is set to be $3$ in all our experiments.
We train the human pose estimation model by minimizing the following loss: 
\begin{equation} \label{eq:loss}
    \mathcal{L}_e = \sum_i^H \sum_j^W \sum_k^K ||\mathbf{H}_{ijk}^{pred} - \mathbf{H}_{ijk}^{gt}||_2^2,
\end{equation}
where $H$ and $W$ represent the height and width of heatmaps, and $K$ is the 
number of joints.

\subsection{Dynamics Modeling via GNN}

\begin{figure}[]
\begin{center}
    \includegraphics[width=0.99\linewidth]{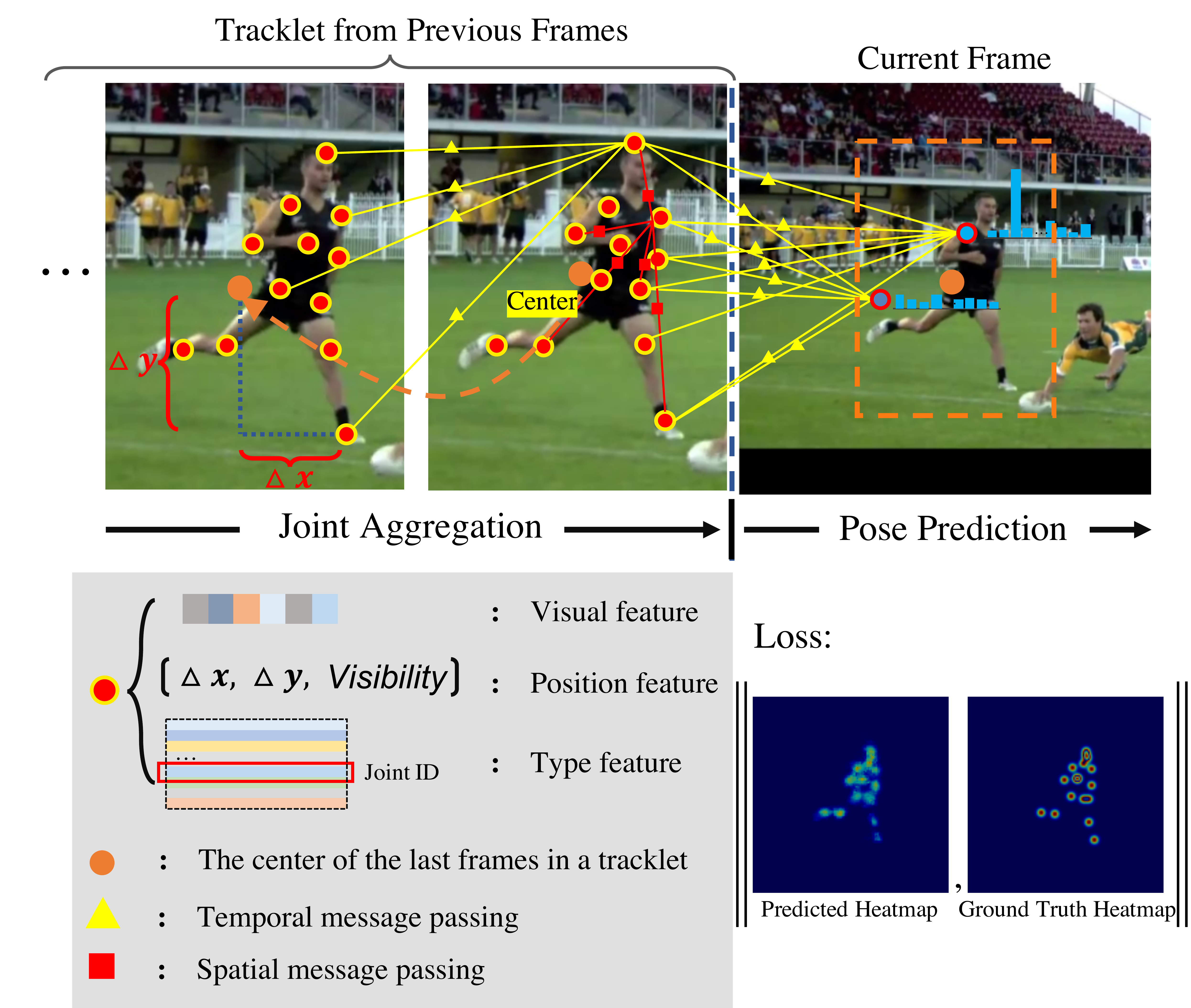}
\end{center}
\vspace{-1em}
  \caption{Illustration of our GNN model. Nodes in the tracklet are the joints of poses, 
  while edges are the connections between joints within the same pose or across consecutive poses. 
  During the pose prediction,
  we model each position in the current frame 
  as a node and generate the heatmaps 
  by classifying all the nodes. $L_2$ norm is used as the loss function to train the GNN model.}
\vspace{-1em}
\label{fig:gnn}
\end{figure}


As shown in Figure~\ref{fig:pipeline}, given the tracked poses of the same identity from prior frames, we design a GNN-based model 
to explicitly capture the spatial-temporal human motion dynamics from history poses and make prediction for the subject's pose in current frame. 

The GNN as a human pose dynamics model has joints of tracklets as the nodes. 
Edges between all pair-wise joints within-frame and between consecutive frames help capture the relative location constraints between joints as well as human motion dynamics. When applied to history tracklets as shown in the Joint Aggregation part in Figure~\ref{fig:gnn}, the GNN updates features on the nodes with respective to the learned dynamics. 
For pose prediction, each location in the current frame is considered as a node and is connected to the joints of the last pose in the tracklet. The GNN performs feature aggregation for the locations in the current frame and classifies each location by its aggregated features to determine the joint type of the location.


Let $t$ be the total number of frames involved in the GNN. A FIFO queue is used
to maintain the history poses with the same identity.
We denote a human pose as $\mathbf{P}_r$, where $r \in \{1,\dots,t\}$.
$\mathbf{P}_{1,\dots,t-1}$ are from history tracklets and $\mathbf{P}_t$ 
represents the predicted pose in the current frame. 



\subsubsection{Nodes in the proposed GNN model}

Joints of history tracklets and potential joints of the human pose in current frame 
are used as nodes in our GNN model. For each frame, we incorporate three kinds of cues on each joint to construct 
the input node feature, 
the visual feature from the backbone CNN of our single-frame pose estimator as $v_k$, 
the encoding of its joint type with a learnable lookup table~\cite{devlin2018bert} as $c_k$, 
and its 2D position and confidence score from pose estimator as $p_k$. 
For the potential joint in the current frame we set its confidence to $1$. All the 2D position of joints are normalized according to the center of the last tracked pose $\mathbf{P}_{t-1}$. 
Normalizing joint positions with respect to the same center help capture the full body movement. 
Here $k\in{1,\dots,K}$ denotes the $k$-th joint type of a given human pose.

We use Multilayer Perceptron~(MLP) to transform all the joint features
to have the same dimension and merge them with average pooling, i.e.
The final feature of the $k$-th joint is computed as follows:
\begin{equation} \label{eq:joints}
\begin{split}
    \mathbf{J}_k = \textbf{Pooling} \big(  & \mathbf{MLP}_{vis}(v_k), \mathbf{MLP}_{pos}(p_k), \\
    & \mathbf{MLP}_{type}(c_k) \big).
\end{split}
\end{equation}

The three $\mathbf{MLP}_{*}$ encoders above ($\mathbf{MLP}_{vis}$, $\mathbf{MLP}_{pos}$, $\mathbf{MLP}_{type}$) for different cues do not share parameters. 
When constructing $\mathbf{J}_k$ for potential joints in the current frame 
the $c_k$ part is ignored.

\subsubsection{Edges in the proposed GNN model}

The graph is constructed with two different types of edges:
the connections between joints within the same frame and
the connections across consecutive frames. Edges within the same frame 
enable the GNN to capture relative movements and spatial structure of human joints 
while the cross-frame edges model the temporal human pose dynamics. 
We use two sets of GNN parameters 
when aggregating features from these two types of edges.

\subsubsection{Joint aggregation}

In each layer of the GNN model, node features
are updated via massage passing, i.e.,
\begin{equation} \label{eq:mp}
   \mathbf{J}_k^{l+1} 
   = {\mathbf{J}_k^{l}} 
   + \mathbf{MLP}\Big( 
            \big[{\mathbf{J}_k^{l}} 
            || \mathbf{M}( {\mathbf{J}_{k^\prime,k^\prime \in \mathcal{N}_{\mathbf{J}_k^l}}^{l} }
            | \mathbf{J}_k^l ) \big]
                \Big),
\end{equation}
where $\mathbf{J}_k^l$ represents the feature of the $k$-th joint at
the $l$-th layer. 
$\mathcal{N}_{\mathbf{J}_k^l}$ represents
the set of neighbours of the $k$-th joint,
$\mathbf{M}$ represents the message aggregating function
that takes all the neighbours as inputs and computes
the aggregated feature, 
and $[\cdot||\cdot]$ represents the concatenation of vectors.

We use self-attention~\cite{vaswani2017attention} mechanism 
in function $\mathbf{M}$ to compute the aggregated feature.
To aggregate the features from all the neighbours,
the query representation of $\mathbf{J_k}$ is computed as $\mathbf{J_{kq}}$
and then each joint $\mathbf{J_{k^\prime}}$ is first transformed to two different 
representations include value $\mathbf{J_{k^\prime v}}$ and key $\mathbf{J_{k^\prime k}}$.
The final aggregated feature can be computed as the 
weighted average of all the values of the neighbours:
\begin{equation} \label{eq:attention}
\begin{split}
    \mathbf{M}( {\mathbf{J}_{k^\prime,k^\prime \in \mathcal{N}_{\mathbf{J}_k}}}
    |\mathbf{J}_k ) = 
    \sum_{k^\prime \in \mathcal{N}_{\mathbf{J}_k} } \alpha_{kk^\prime} \mathbf{J}_{k^\prime v}, \\
    \text{where } \alpha_{kk^\prime} = \mathbf{Softmax}_{k^\prime} (\mathbf{J}_{kq} ^\top \mathbf{J}_{k^\prime k}).
\end{split}
\end{equation}

$\mathbf{J}^\top$ represents the transpose of the feature vector $\mathbf{J}$ 
and the similarity is computed as the dot product between the query and keys.
$\alpha_{kk^\prime}$ is computed as the softmax normalization over the 
similarities.

The information comes from the different types of edges plays different roles: edges within the same frame model the spatial dynamics while
edges across frames incorporate the temporal dynamics. 
We keep separated parameters for the two dynamics. Specifically, the $\mathbf{MLP}(\boldsymbol{\cdot})$ in Equation~\ref{eq:mp} is switched between two implementations from layer to layer. In the $l$-th layer, the implementation is set to be $\mathbf{MLP}_{spatial}(\boldsymbol{\cdot})$ working on neighbors defined by the edges within the same frame and in the next $(l+1)$-th layer, it is switched to $\mathbf{MLP}_{temporal}(\boldsymbol{\cdot})$ working on neighbors defined by edges across frames, and so on so forth.
The aggregated features of joints from $\mathbf{P}_{t-1}$ are used for the pose prediction step.

\subsubsection{Pose prediction~\label{subsub:prediction}}

This step aims to locate the poses in current frame by the GNN model, with neither human detection nor single-frame human pose estimator. 
To reduce computation, we select potential joints only from a confined scope. 
We propagate the bounding box of the last tracked pose $\mathbf{P}_{t-1}$ 
to current frame and scale it up by a factor of $1.5$ 
vertically and $2$ horizontally at the same center to support fast-motion scenes, shown as the dotted orange box in
Figure~\ref{fig:gnn}.

A graph is constructed with potential joints in the current frame and joints from $\mathbf{P}_{t-1}$. The learned GNN model is then applied to this graph to update joint features via message passing as explained above.

On top of the final features from GNN as $\mathbf{J}$, the prediction is conducted via another MLP over each potential joint in current frame, i.e.,
\begin{equation}
    \mathbf{Prob} = \mathbf{MLP}_{pred}(\mathbf{J}),
\end{equation}
where $\mathbf{Prob}$ denotes the probability distribution over all joint types of the input node. The predicted probability distributions of all potential joints in current frame generate the predicted heatmaps for all joints.

\subsubsection{Training}

As in Equation~\ref{eq:loss}, we generate ground-truth heatmaps from labeled human pose and compute $L_2$ loss against the predicted joint heatmaps. 
Since the full GNN predictor is differentiable, we optimize the parameters and learn the dynamics from end to end.



\subsection{Online Tracking Pipeline} 

In the current frame, given the poses from the GNN-based predictor and the poses
from the single-frame pose estimator, we match and fuse them to obtain the final tracked human poses. 
In this process, the poses from the predictor and that from the estimator 
are complimentary to each other as the poses missed by the single-frame estimator due to occlusion and motion blur can be recovered by the predictor. 


Specifically, we apply Hungarian matching to compute an one-to-one mapping 
between the predicted poses and the estimated poses. 
The similarity used in the Hungarian algorithm is the 
object keypoint similarity~\cite{FlowTrack} computed based on the positions of the joints.

After matching, we propagate the tracking IDs from the predicted poses to
the estimated poses if they are matched.
A new ID is assigned to the estimated pose without a matched predicted pose, 
which is likely to be a newly observed one.
For all the matched poses, the joint heatmaps of the two poses are
first aligned according to their centers and then
merged together by averaging the heatmaps. Refined poses are then decoded from the fused heatmaps.

We store the tracked results in a FIFO manner 
while keeping a fixed size of each tracklet. 
The history tracklets are then used as inputs to the GNN model 
for the following frame. The proposed framework is hence implemented to be an online tracker, as shown in Figure~\ref{fig:pipeline}.

\section{Experiments}

\subsection{Datasets}

We evaluate the proposed method on two widely used datasets for human
pose estimation and tracking, PoseTrack 2017 
and PoseTrack 2018~\cite{ArtTrack}.
These datasets contain several video sequences of articulated people
that perform various actions. 
Specially, PoseTrack 2017 contains 250 video sequences for training
and 50 video sequences for validation, 
PoseTrack 2018 increases the number of video sequences 
and contains 593 for training and 170 for validations.
Both datasets are annotated with 15 joints, each of them
are associated with an ID for the corresponding person.
The training videos are annotated densely within the middle 30 frames 
of each video sequences.
The validation videos are annotated every forth frame across the whole
video sequences beside the densely annotation of the middle.
We use the training set for training and validation set for testing,
which is a common setup in previous works~\cite{TKMRNet,DAT}.

The performance of the proposed method is evaluated from two
aspects: human pose estimation and human pose tracking. 
We use mean Average Precision (mAP)~\cite{lin2014microsoft_map,ruggero2017benchmarking_map} to evaluate the performance of human pose estimation, and Multi Object Tracking Accuracy (MOTA) to evaluate human pose tracking.
MOTA is evaluated based on three kinds of errors:
missing rate, false positive rate, and switch rate.
Both metrics are computed independently for
each joint and then averaged across all joints.
Since the evaluation of human pose tracking requires
filtering the joints according to some certain thresholds,
we can either evaluated the performance of 
human pose estimation independently or based on
the filtered joints.
The former one provides us an illustration of
of the trade-off between human pose tracking and 
human pose estimation while the
latter one provides us the pure performance of human
pose estimation.
We report both results for pose estimation.

\subsection{Implementation Details}

For the single-frame human pose estimation, we used HRNet~\cite{sun2019deepHRNet} 
as the backbone. 
Following the training strategies of~\cite{PoseWarper,TKMRNet},
the HRNet is first trained on COCO dataset and then fine-tuned on
PoseTrack 2017 and PoseTrack 2018 independently.
For the fine-tuning process, we train the model for 20 epochs with Adam
optimizer. The initial learning rate is set to be 0.0001 
and reduced by a factor of 10 at the 10th and 15th epochs. 
We add several data augmentation strategies as used in~\cite{PoseWarper},
including random rotation, random flip, 
randomly using half of body, and random scale.
Flip test is used in our work as in~\cite{CombDet}.
We adopt Faster R-CNN~\cite{ren2016faster} with feature pyramid network and 
deformable convolutional network as the human detector~\cite{TKMRNet}.
The human detector is pre-trained on COCO dataset and then fine-tuned 
on PoseTrack 2017 and PoseTrack 2018 separately. 

For the human detector, Non-Maximum Suppresion (NMS) is applied
to remove duplicate detected bounding boxes which is a common
operation in detection. 
Specifically, we use Soft-NMS~\cite{bodla2017soft} and set the threshold
to 0.7. As articulated human pose tracking in a video
often involves complex interaction and heavy person-to-person occlusions, traditional NMS in object detection that merely rely
on the Intersection Over Union~(IOU) of the bounding boxes is prone to
fail~\cite{TKMRNet}.
Since we have the pose information, Pose-based Non-Maximum
Suppresion~(pNMS)~\cite{fang2017rmpe_pnms} is adopted to help
further remove the duplicate human poses.
In pNMS, the IOU is not computed based on the bounding boxes
but the weighted sum of all the joints' distances with respect
to the scale of the pose. The threshold of pNMS is set to be 
0.5.

For the training of the GNN pose prediction model,
the fine-tuned backbone model is used to compute the visual feature
of the joints. Specifically, we obtain the feature maps that are in the same
resolution as the heatmaps, from all the three stages
of the HRNet. The feature maps then are concatenated together and form
the final feature maps with depth of 144. 
The visual feature of each joint can be obtained according to the joint position in the heatmap.
Several data augmentation strategies are used during the GNN training process, including random rotation of the tube, random flip,
random scale of the tube, and 
randomly selecting the gap between consecutive frames in the tube.
We train the GNN model for 10 epochs with Adam
optimizer. The initial learning rate is set to be 0.0001 
and reduced by a factor of 10 at the 5th and 8th epochs. The length of pose history is set to be three.

\subsection{Results on PoseTrack 2017}

We compare our proposed method with the state-of-the-art methods
in human pose estimation and human pose tracking,
which are shown in Table~\ref{tab:sota_pure_map_17}, Table~\ref{tab:sota_map_17}, and Table~\ref{tab:sota_mota_17}. 
In Table~\ref{tab:sota_map_17} and Table~\ref{tab:sota_mota_17}, 
the upper methods are bottom-up fashion and lower methods 
are top-down fashion.

\begin{table}[b]
    \centering
    \vspace{-1em}
    \scalebox{0.71}{
    \begin{tabular}{c | c c c c c c c | c}
    \hline
        Method & Head & Shou & Elb & Wri & Hip & Knee & Ankl & Total \\
    \hline
    PoseWarper~\cite{PoseWarper} & 81.4 &  88.3 & 83.9 & 78.0 & 82.4 & 80.5 & 73.6 & 81.2 \\ 
    CombDet~\cite{CombDet}  & 89.4 &   89.7 & 85.5 & \textbf{79.5}  & 82.4 & 80.8 &  76.4 & 83.8 \\ 
    Ours    & \textbf{90.9} & \textbf{90.7} & \textbf{86.0} & 79.2 & \textbf{83.8} & \textbf{82.7} & \textbf{78.0} & \textbf{84.9}  \\ 
    \hline    
    \end{tabular}
    }
    \caption{Comparison of state-of-the-art methods on pure human pose estimation (without filtering) on the PoseTrack 2017 validation set, where the
    performance is evaluated as mAP and all joints are counted.}
    \label{tab:sota_pure_map_17}
\end{table}

 \textbf{Human pose estimation.}
In Table~\ref{tab:sota_pure_map_17} and Table~\ref{tab:sota_map_17}, we evaluate pure human
pose estimation in videos where the estimated poses are directly evaluated without filtering, as well as the filtered human pose estimation performance in the context of pose tracking. As shown in Table~\ref{tab:sota_pure_map_17}, the proposed method achieves the 
best performance, outperforming the previous best method~\cite{CombDet} by 1.1 mAP.
Note that CombDet~\cite{CombDet} utilizes a heavier 3D convolutional backbone and uses 9 frames as input. Since human pose tracking needs to firstly filter some estimated joints, the mAP result in Table~\ref{tab:sota_map_17} is lower than that in Table~\ref{tab:sota_pure_map_17}. As shown in Table~\ref{tab:sota_mota_17}, our method outperforms the best top-down method~\cite{TKMRNet} by 1.6 mAP and the best bottom-up method~\cite{ST-Embed} by
4.1 mAP.

\textbf{Human pose tracking.}
As shown in Table~\ref{tab:sota_mota_17}, our method achieves state-of-the-art pose tracking
performance and outperform the best top-down method~\cite{TKMRNet} by 1.2 MOTA, and the best bottom-up method~\cite{ST-Embed} by 1.6 MOTA. 

\begin{figure*}[t]
\begin{center}
    \includegraphics[width=0.9\linewidth]{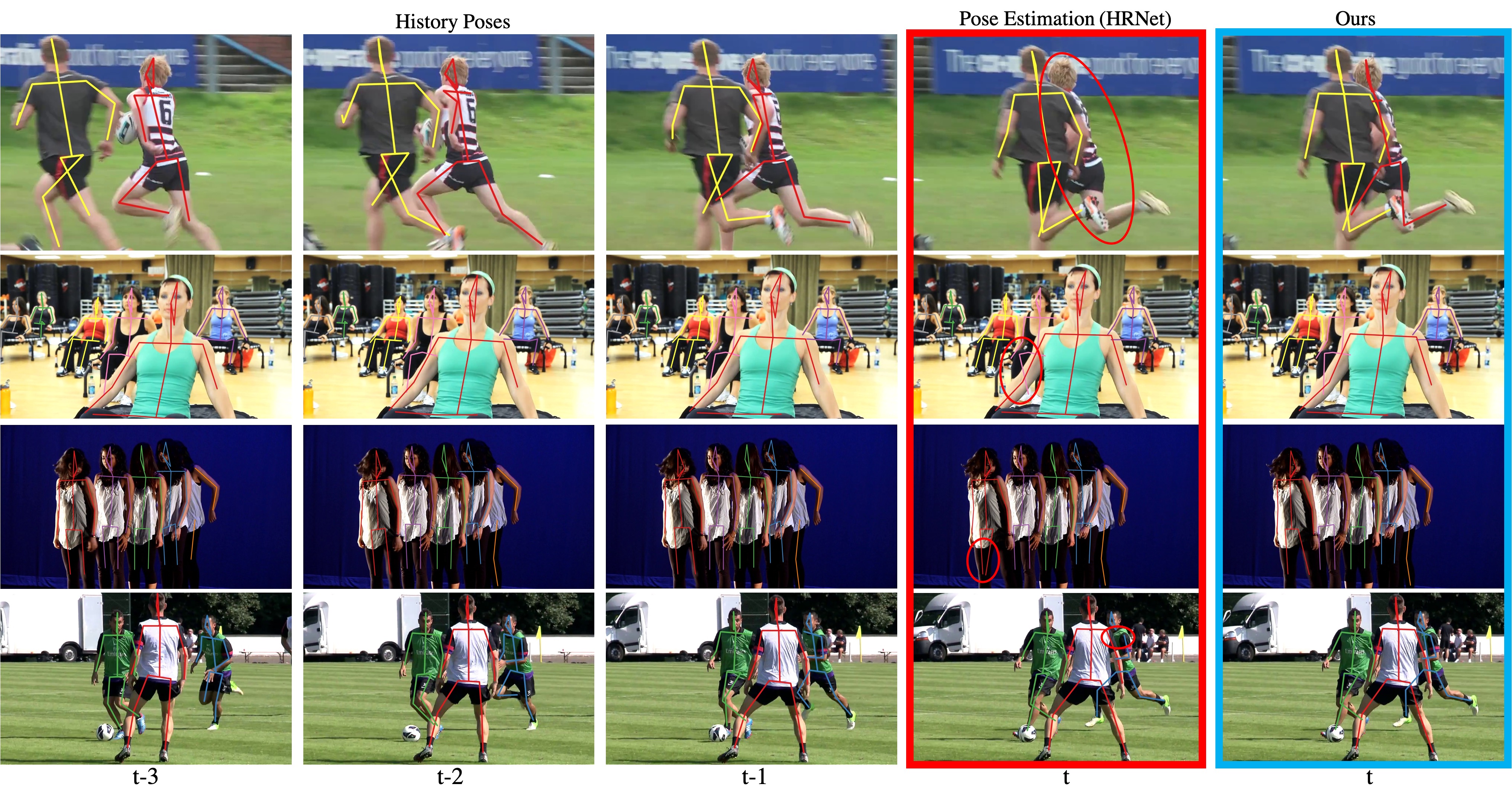}
\end{center}
\vspace{-1.6em}
  \caption{Qualitative examples of the proposed method on
  the PoseTrack 2017 validation set. The first three columns show
  the poses in the memory, the fourth column shows the estimated poses 
  from HRNet, and the last column shows the final poses of our proposed method.
  Red dot circles highlight the incorrect or missed poses that are corrected.}
\vspace{-1em}
\label{fig:samples}
\end{figure*}

\textbf{Qualitative samples.}
To provide an intuitive understanding of our method, 
in Figure~\ref{fig:samples} we visualize some samples of the
history pose tracklets, the pose estimation result in the current frame, and 
the final outputs of our full method. Different skeleton colors represents different person identity
and the red circles in the 4th column highlight the missed or incorrect estimated joints that are corrected by the proposed GNN model.

\begin{table}[]
    \centering
    \scalebox{0.71}{
    \begin{tabular}{c | c c c c c c c | c}
    \hline
        Method & Head & Shou & Elb & Wri & Hip & Knee & Ankl & Total \\
    \hline
    BUTD~\cite{BUTD} & 79.1 & 77.3 & 69.9 & 58.3 & 66.2 & 63.5 & 54.9 & 67.8 \\
    RPAF~\cite{RPAF}       & 83.8 &  84.9 & 76.2 & 64.0 & 72.2 & 64.5 & 56.6 & 72.6 \\ 
    ArtTrack~\cite{ArtTrack}   & 78.7 &  76.2 & 70.4 & 62.3 & 68.1 & 66.7 & 58.4 & 68.7 \\ 
    PoseFlow~\cite{PoseFlow}   & 66.7 &  73.3 & 68.3 & 61.1 & 67.5 & 67.0 & 61.3 & 66.5 \\ 
    STAF~\cite{STAF} & -    &  -    &  -   & 65.0 &  -   & -    & 62.7 & 72.6 \\ 
    ST-Embed~\cite{ST-Embed}   & 83.8 &  81.6 & 77.1 & 70.0 & 77.4 & 74.5 & 70.8 & 77.0 \\ 
    \hline
    DAT~\cite{DAT}        & 67.5 &  70.2 & 62.0 & 51.7 & 60.7 & 58.7 & 49.8 & 60.6 \\ 
    FlowTrack~\cite{FlowTrack}  & 81.7 &  83.4 & 80.0 & 72.4 & 75.3 & 74.8 & 67.1 & 76.9 \\ 
    TKMRNet~\cite{TKMRNet}    & 85.3 &  88.2 & 79.5 & 71.6 & 76.9 & 76.9 & \textbf{73.1} & 79.5 \\ 
    Ours       & \textbf{88.4} & \textbf{88.4} & \textbf{82.0} & \textbf{74.5} & \textbf{79.1} & \textbf{78.3} & \textbf{73.1} & \textbf{81.1}   \\ 
    \hline    
    \end{tabular}
    }
    \caption{Comparison with state-of-the-art methods on human pose estimation (with filtering) on the
    PoseTrack 2017 Validation set, where thresholds are used to filtering 
    low confidence joints for pose tracking. Evaluated in mAP and all joints are counted. }
    \label{tab:sota_map_17}
\end{table}

\begin{table}[]
    \centering
    \scalebox{0.71}{
    \begin{tabular}{c | c c c c c c c | c}
    \hline
    Method     & Head & Shou & Elb  & Wri  & Hip  & Knee & Ankl & Total \\
    \hline
    BUTD~\cite{BUTD}       & 71.5 & 70.3 & 56.3 & 45.1 & 55.5 & 50.8 & 37.5 & 56.4 \\
    ArtTrack~\cite{ArtTrack}   & 66.2 & 64.2 & 53.2 & 43.7 & 53.0 & 51.6 & 41.7 & 53.4 \\
    PoseFlow~\cite{PoseFlow}   & 59.8 & 67.0 & 59.8 & 51.6 & 60.0 & 58.4 & 50.5 & 58.3 \\
    STAF~\cite{STAF}       & -    & -    & -    & -    & -    & -    & -    & 62.7 \\
    ST-Embed~\cite{ST-Embed}   & 78.7 & 79.2 & 71.2 & 61.1 & \textbf{74.5} & 69.7 & \textbf{64.5} & 71.8 \\
    \hline
    DAT~\cite{DAT}        & 61.7 & 65.5 & 57.3 & 45.7 & 54.3 & 53.1 & 45.7 & 55.2 \\
    FlowTrack~\cite{FlowTrack}  & 73.9 & 75.9 & 63.7 & 56.1 & 65.5 & 65.1 & 53.5 & 65.4 \\
    PGPT~\cite{bao2020poseGuided} & 75.4 & 77.2 & 69.4 & 71.5 & 65.8 & 67.2 & 59.0 & 68.4 \\
    TKMRNet~\cite{TKMRNet}    & 81.0 & 82.9 & 69.8 & 63.6 & 72.0 & 71.1 & 60.8 & 72.2 \\
    CombDet~\cite{CombDet}    & 80.5 & 80.9 & 71.6 & \textbf{63.8} & 70.1 & 68.2 & 62.0 & 71.6 \\
    Ours       & \textbf{82.0} & \textbf{83.1} & \textbf{73.4} & 63.5 & 72.3 & \textbf{71.3} & 63.5 & \textbf{73.4} \\
    \hline
    \end{tabular}
    }
    \caption{Comparison of state-of-the-art methods on human pose tracking on the PoseTrack 2017 validation set. The performance is evaluated as MOTA and all joints are counted.}
    \vspace{-1em}
    \label{tab:sota_mota_17}
\end{table}

\subsection{Results on PoseTrack 2018}

We show in Table~\ref{tab:sota_pure_map_18}, Table~\ref{tab:sota_map_18} 
and Table~\ref{tab:sota_mota_18} the comparison of our proposed method
and existing methods on the PoseTrack 2018
validation set. Again, our method achieves the best performances
in pure human pose estimation, pose estimation with filtering, and human pose tracking.
Specifically, as in Table~\ref{tab:sota_pure_map_18}, the proposed method improves pure human pose estimation without filtering by 1.1 mAP over the state-of-the-art method~\cite{KeyTrack}. As shown in Table~\ref{tab:sota_map_18}, our method outperforms the best existing human pose estimation~\cite{TKMRNet} with filtering for pose tracking by 1.2 mAP.
And for human pose tracking, the proposed method
also achieves the state-of-the-art performance improving the MOTA by 0.3 over~\cite{TKMRNet}, as shown in Table~\ref{tab:sota_mota_18}.

The superior performance on both PoseTrack 2017 and 2018 datasets in all three tasks (pure pose estimation in video, pose estimation with filtering, and pose tracking) validates the effectiveness of modeling dynamics by GNN.

\begin{table}[]
    \centering
    \scalebox{0.71}{
    \begin{tabular}{c | c c c c c c c | c}
    \hline
        Method & Head & Shou & Elb & Wri & Hip & Knee & Ankl & Total \\
    \hline
    PT\_CPN++~\cite{PT_CPN++}    & 82.4 & 88.8 & 86.2 & 79.4 & 72.0 & 80.6 & 76.2 & 80.9 \\
    KeyTrack~\cite{KeyTrack}  & 84.1 & 87.2 & \textbf{85.3} & 79.2 & 77.1 & 80.6 & 76.5 &   81.6 \\
    CombDet~\cite{CombDet} & 84.9 & 87.4 & 84.8 & 79.2 & 77.6 & 79.7 & 75.3 & 81.5 \\
    Ours   & \textbf{85.1} & \textbf{87.7} & \textbf{85.3} & \textbf{80.0} & \textbf{81.1} & \textbf{81.6} & \textbf{77.2} & \textbf{82.7}   \\
    \hline
    \end{tabular}
    }
    \caption{Comparison of state-of-the-art methods on pure human pose estimation (without filtering)
    on the validation set of PoseTrack 2018. Evaluated in mAP and all joints are counted.}
    \label{tab:sota_pure_map_18}
\end{table}

\begin{table}[]
    \centering
    \scalebox{0.71}{
    \begin{tabular}{c | c c c c c c c | c}
    \hline
        Method & Head & Shou & Elb & Wri & Hip & Knee & Ankl & Total \\
    \hline
    STAF~\cite{STAF}        & -    & -    & -    & 64.7 & -    & -    & 62.0 & 70.4 \\
    TML++~\cite{TML++}       & -    & -    & -    & -    & -    & -    & -    & 74.6 \\
    \hline
    TKMRNet~\cite{TKMRNet}     & -    & -    & -    & -    & -    & -    & -    & 76.7 \\
    Ours                    & 80.6 & 84.5 & 80.6 & 74.4 & 75.0 & 76.7 & 71.9 & \textbf{77.9}   \\
    \hline
    \end{tabular}
    }
    \caption{Comparison of state-of-the-art methods on human pose estimation (with filtering)
    on the PoseTrack 2018 validation set, where thresholds are used to filtering 
    low confidence joints for pose tracking.}
    \label{tab:sota_map_18}
\end{table}

\begin{table}[]
    \centering
    \scalebox{0.71}{
    \begin{tabular}{c | c c c c c c c | c}
    \hline
    Method     & Head & Shou & Elb  & Wri  & Hip  & Knee & Ankl & Total \\
    \hline
    STAF~\cite{STAF}       & -    & -    & -    & -    & -    & -    & -    & 60.9 \\
    TML++~\cite{TML++}     & \textbf{76.0} & 76.9 & 66.1 & 56.4 & 65.1 & 61.6 & 52.4 & 65.7 \\
    \hline
    PT\_CPN++~\cite{PT_CPN++}   & 68.8 & 73.5 & 65.6 & 61.2 & 54.9 & 64.6 & 56.7 & 64.0 \\ 
    TKMRNet~\cite{TKMRNet}   & -    & -    & -    & -    & -    & -    & -    & 68.9 \\ 
    KeyTrack~\cite{KeyTrack} & -    & -    & -    & -    & -    & -    & -    & 66.6 \\ 
    CombDet~\cite{CombDet}   & 74.2 & 76.4 & 71.2 & 64.1 & 64.5 & 65.8 & \textbf{61.9} & 68.7 \\
    Ours                     & 74.3 & \textbf{77.6} & \textbf{71.4} & \textbf{64.3} & \textbf{65.6} & \textbf{66.7} & 61.7 & \textbf{69.2} \\
    \hline
    \end{tabular}
    }
    \caption{Comparison of state-of-the-art methods on human pose tracking on the PoseTrack 2018 validation set. Evaluated in MOTA and all joints are counted.}
    \label{tab:sota_mota_18}
    \vspace{-1em}
\end{table}

\subsection{Model Analysis}

We provide here analyses on the proposed method, 
including  ablation
studies, visualization of the attentions among
joints learnt from the GNN model,
and  sensitively
analysis of the memory length and GNN model size.

\begin{figure}[]
\begin{center}
\scalebox{0.9}{
    \includegraphics[width=1.0\linewidth]{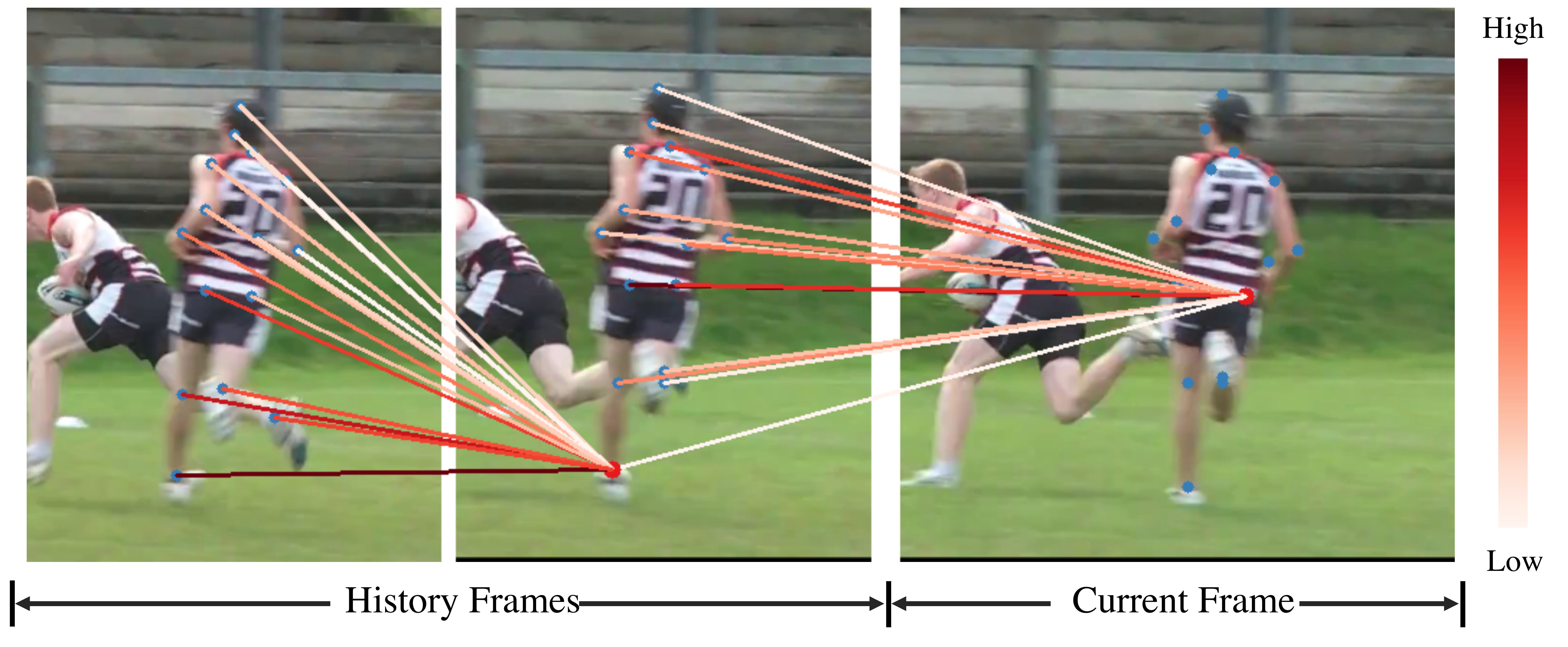}
    }
\end{center}
\vspace{-1em}
  \caption{Visualization of the attention among different joints within
  the GNN model.
  Red nodes are the centers for aggregation and the colors of lines
  indicate the attention values. We zoom the current frame for a better visualization.}
\vspace{-1.5em}
\label{fig:attention_vis}
\end{figure}

\textbf{Ablation study.} We examine the effectiveness of the proposed method by conducting ablation experiments on several key components. As shown in Table.\ref{tab:ablation}, 
Matching w/ IOU and Matching w/ OKS means we 
associate the estimated poses between consecutive frames
using the IOU and OKS as the similarity measure. Matching w/ GNN means we only use the predicted poses for matching measure, and the final poses are not refined by the predicted poses.
Full model is our proposed model.
It can be seen that using the predicted poses for matching metric
can improve the MOTA performance and reduce the switch rate over IOU and OKS metrics, 
the full model with pose refinement by pose merging can improve both the mAP and MOTA
further more.

\begin{table}[b]
    \centering
    \vspace{-1.5em}
    \scalebox{0.76}{
    \begin{tabular}{c | c c | c c c }
    \hline
    Method     & mAP & MOTA & Miss~(\%)  & Switch~(\%)  &  FP~(\%)  \\
    \hline
    Matching w/ IOU          & 79.9 & 71.8 & 17.4 & 1.8 & 9.0 \\ 
    Matching w/ OKS          & 79.9 & 72.1 & 17.4 & 1.6 & 8.9 \\ 
    Matching w/ GNN          & 79.9 & 73.1 & 17.1 & 1.4 & 8.4 \\ 
    Full model               & 81.1 & 73.4 & 16.9 & 1.3 & 8.4 \\ 
    \hline
    \end{tabular}
    }
    \caption{Ablation studies on the PoseTrack 2017 validation set, where Miss, Switch, FP stand for the missing rate, switch rate and false positive rate (the lower the better) in MOTA.}
    \label{tab:ablation}
\end{table} 

\textbf{Visualization of GNN model.}
In order to provide a thorough understanding of the GNN model, we visualize in Figure~\ref{fig:attention_vis} the computed attention weights $\alpha_{kk^\prime}$
as computed in Equation~\ref{eq:attention}.
It can be observed that the hip in current frame is influenced 
by the hip, shoulder, and knee in the consecutive pose mostly.
The ankle in the middle frame is influence mostly by the lower part of the previous pose.

\textbf{Length of memory tube and model size.}
In Table~\ref{tab:modelsize} we show the results with
different lengths of memory and different model size,
where smaller model means the dimension of the output of 
$\mathbf{MLP}_*$ (as in Equation~\ref{eq:joints}) is halved.
It can be seen that the performance is improved when changing the 
memory length from two to four frames and being saturated when using more memory.
Enlarging the model size improves both mAP and MOTA.

\begin{table}[]
    \centering
    \scalebox{0.77}{
    \begin{tabular}{c | c c c c c }
    \hline
    Method     & mAP & MOTA & Miss~(\%)  & Switch~(\%)  &  FP~(\%)  \\
    \hline
    Two frames             & 80.6 & 72.9 & 17.2 & 1.4 & 8.5  \\ 
    Four frames              & 81.3 & 73.4 & 16.9 & 1.3 & 8.4 \\ 
    \hline
    Smaller model            & 80.8 & 73.2 & 17.1 & 1.3 & 8.4 \\ 
    \hline
    Full model               & 81.1 & 73.4 & 16.9 & 1.3 & 8.4 \\ 
    \hline
    \end{tabular}
    }
    \caption{Influence of model capacity and length of memory.}
    \label{tab:modelsize}
    \vspace{-1.0em}
\end{table} 


\section{Conclusion}
We present in this paper a novel 
approach for human pose estimation and tracking.
In our method, a GNN model is designed to
explicitly model the dynamics of the pose tracklets and predict 
the corresponding poses in an incoming frame,
independent of the estimations.
When combining with the human pose estimation model, 
the proposed method 
takes advantages of 
both the visual information
and the dynamics, thereby 
enabling the recovery of missed poses
and refinement of estimated poses.
Extensive experiments on PoseTrack 
2017 and PoseTrack 2018 datasets
validate the superiority of the proposed method in both
human pose estimation and human pose tracking tasks.
In our future work, we would like to explore
a more flexible manner to aggregate the 
predicted results and the new observation,
making the whole pipeline even more adaptive.


{\small
\bibliographystyle{ieee_fullname}
\bibliography{references}
}

\end{document}